\documentclass[nohyperref]{article}

\usepackage{microtype}
\usepackage{graphicx}
\usepackage{booktabs} 
\usepackage{graphicx}
\usepackage{amsmath}
\usepackage{amssymb}
\usepackage{booktabs}

\usepackage{times}
\usepackage{epsfig}
\usepackage{graphicx}
\usepackage{amsmath}
\usepackage{amssymb}
\usepackage{multirow}
\usepackage{subfloat}
\usepackage{subfig}

\usepackage{booktabs}
\usepackage{tabulary}
\usepackage[dvipsnames]{xcolor}
\usepackage{soul}
\usepackage{xspace}\xspace
\usepackage{adjustbox}
\usepackage{array}

\usepackage{dblfloatfix}
\usepackage{transparent}

\usepackage{algorithm}
\usepackage{listings}

\newcolumntype{x}[1]{>{\centering\arraybackslash}p{#1pt}}

\newcommand{\app}{\raise.17ex\hbox{$\scriptstyle\sim$}}

\def\x{$\times$}
\newcolumntype{x}[1]{>{\centering\arraybackslash}p{#1pt}}

\newlength\savewidth\newcommand\shline{\noalign{\global\savewidth\arrayrulewidth
  \global\arrayrulewidth 1pt}\hline\noalign{\global\arrayrulewidth\savewidth}}
\newcommand{\tablestyle}[2]{\setlength{\tabcolsep}{#1}\renewcommand{\arraystretch}{#2}\centering\footnotesize}

\usepackage{hyperref}

\usepackage[accepted]{icml2022}

\usepackage{amsmath}
\usepackage{amssymb}
\usepackage{mathtools}
\usepackage{amsthm}

\usepackage[capitalize,noabbrev]{cleveref}

\theoremstyle{plain}

\theoremstyle{definition}

\theoremstyle{remark}

\usepackage[textsize=tiny]{todonotes}

\icmltitlerunning{witcherofresearch@gmail.com}
\begin{document}

\twocolumn[
\icmltitle{TFCNet: Temporal Fully Connected Networks for static unbiased temporal reasoning}




\begin{icmlauthorlist}

\icmlauthor{Shiwen Zhang}{comp}

\end{icmlauthorlist}

\icmlaffiliation{comp}{witcherofresearch@gmail.com}

\icmlcorrespondingauthor{Shiwen Zhang}{}

\icmlkeywords{Machine Learning, ICML}

\vskip 0.3in
]



\begin{abstract}
Temporal Reasoning is one important functionality for vision intelligence. In computer vision research community, temporal reasoning is usually studied in the form of video classification, for which many state-of-the-art Neural Network structures and dataset benchmarks are proposed  in recent years, especially 3D CNNs and Kinetics. However, some recent works found that current video classification benchmarks contain strong biases towards static features, thus cannot accurately reflect the temporal modeling ability. New video classification benchmarks aiming to eliminate static biases are proposed, with experiments on these new benchmarks showing that the current clip-based 3D CNNs are outperformed by RNN structures and recent video transformers.

    In this paper, we find that 3D CNNs and their efficient depthwise variants, when video-level sampling strategy is used, are actually able to beat RNNs and recent vision transformers by significant margins on static-unbiased temporal reasoning benchmarks. Further, we propose Temporal Fully Connected Block (TFC Block), an efficient and effective component, which approximates fully connected layers along temporal dimension to obtain video-level receptive field, enhancing the spatiotemporal reasoning ability. With TFC blocks inserted into Video-level 3D CNNs (V3D), our proposed TFCNets establish new state-of-the-art results on synthetic temporal reasoning benchmark, CATER, and real world static-unbiased dataset, Diving48, surpassing all previous methods. 

\end{abstract}
\section{Introduction}
Temporal Reasoning in videos requires the computer vision models to understand complicated spatiotemporal patterns and reason among frames, which is often studied in the form of video classification. Many state-of-the-art spatiotemporal learning models have been proposed in recent years, especially 3D CNNs\cite{DBLP:conf/cvpr/CarreiraZ17,DBLP:conf/cvpr/WangL0G18,DBLP:conf/cvpr/0004GGH18,DBLP:journals/corr/abs-1812-03982}, which have achieved state-of-the-art accuracy on mainstream benchmarks, Kinetics-400, UCF-101 etc\cite{DBLP:conf/cvpr/CarreiraZ17,DBLP:journals/corr/abs-1212-0402}. However, some recent works\cite{Girdhar2020CATER:,Li_2018_ECCV} demonstrate that many mainstream benchmarks have strong biases towards static representations, for example, scene and objects, which means that many categories can be recognized by considering only spatial features, without considering temporal features.  In fact, there is no problem with that for video action recognition, since many data in the wild inevitably lead to biases. However, for the purpose of benchmarking the model's ability for temporal reasoning, such 2D spatial biases should be controlled or eliminated. According to \cite{Li_2018_ECCV}, if a benchmark contains representation biases, a model which learns better such biases leads to higher accuracy on this benchmark. Thus if a video benchmark contains strong static biases and temporal biases, it is ambiguous to explain the performance of models on this benchmark.

\begin{figure}[t]
  \centering
   \includegraphics[width=0.8\linewidth]{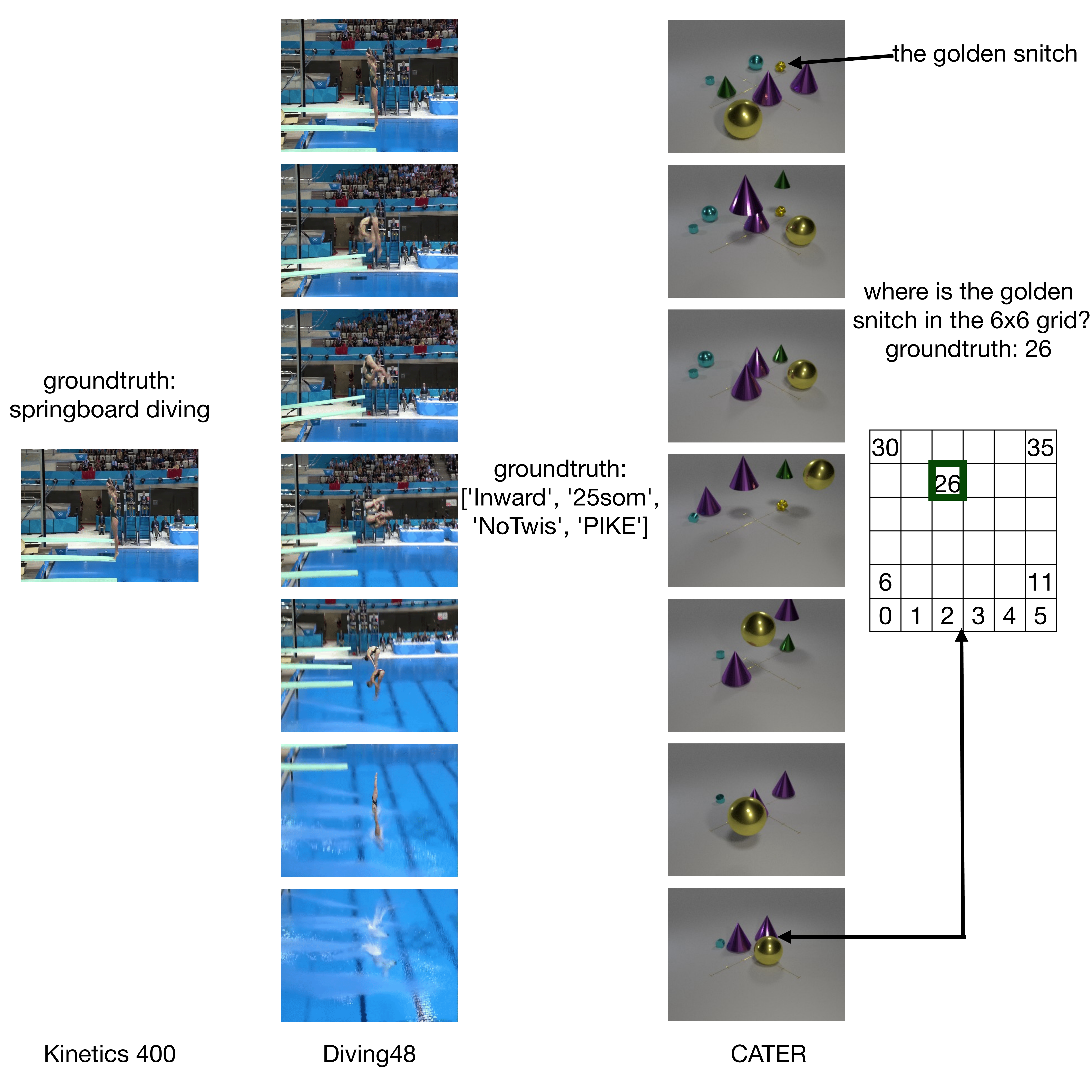}

   \caption{comparison of static-biased Kinetics-400 and static-unbiased video benchmarks Diving48 and CATER.
   }
   \label{fig:intro}
\end{figure}

In \cref{fig:intro}, we show cases of static-biased benchmarks and static-unbiased benchmarks, from static-biased video classification benchmark Kinetics-400\cite{DBLP:conf/cvpr/CarreiraZ17}, synthetic static-unbiased object permanence benchmark CATER\cite{Girdhar2020CATER:}, and real world static-unbiased temporal reasoning benchmark Diving48\cite{Li_2018_ECCV}. In the first column, even with one frame, it is still easy to recognize that the label should be "springboard diving" since the static features are enough to discriminate the label. However, for Diving48, which contains 48 fine-grained diving action labels, it is impossible to distinguish the label with one frame. Also, for CATER snitch localization task, there is a golden snitch appearing in the first frame, which can later be contained, moved and uncontained by other objects in the video. The task asks the model to reason where is the golden snitch in the last frame. Please note that the snitch can be visible, contained or occluded in the last frame, which requires the model to reason the whole video and even state-of-the-art tracking models fail\cite{Girdhar2020CATER:}. The properties of such static-unbiased benchmarks\cite{Girdhar2020CATER:,Li_2018_ECCV} are so different from previous static-biased benchmarks\cite{DBLP:conf/cvpr/CarreiraZ17,DBLP:journals/corr/abs-1212-0402} that it has been shown many state-of-the-art clip-level 3D CNNs\cite{DBLP:journals/corr/abs-1812-03982} fail on these benchmarks\cite{Girdhar2020CATER:,zhou2021hopper} and also are surpassed by long range models, for example LSTM\cite{Hochreiter1997LongSM} and video transformers\cite{bertasius2021spacetime}. Thus it is an important direction to explore the properties of 3D CNNs on these static-unbiased video benchmarks.

 Our contributions  in this paper are from three aspects. First, we found that when video-level sampling strategy is used, 3D CNNs and our proposed efficient depthwise variants , which we term {\bf V3D}, surpass all LSTM and video transformers by significant margins in terms of temporal reasoning capabilities. Second, we propose novel Temporal Fully Connected Operations (TFC Operations), which efficiently obtain global temporal receptive field with few extra parameters and cheap computation. Finally, we integrate the proposed TFC Operations with V3D to form TFCNets, a new type of  video-level temporal reasoning models, which achieve new state-of-the-art results on static-unbiased temporal reasoning benchmarks, CATER and Diving48.

\section{Related Work}

{\bf Video Classification Models} In deep learning era, there are mainly 3 types of network structure for video classification. First, Two-stream convolutional networks were originally proposed by \cite{DBLP:conf/nips/SimonyanZ14}, where one stream is used for learning from RGB images, and the other one for optical flow, with late fusion yielding the final prediction. Although the accuracy of Two-stream networks is high,  the main limitation of such structures is that the computation of optical flow is highly expensive.
Second, various 3D CNNs have  been proposed  \cite{DBLP:conf/iccv/TranBFTP15,DBLP:conf/cvpr/CarreiraZ17,DBLP:conf/cvpr/WangL0G18,DBLP:conf/cvpr/0004GGH18,DBLP:journals/corr/abs-1812-03982}, where 3D convolutions or temporal convolutions plus spatial convolutions are applied directly on video sequence to model spatiotemporal features. It is noteworthy that most of 3D CNNs are clip-based methods, which only sample a random clip of the whole video during training.
Third, long-term modeling frameworks have been developed for capturing long range temporal structures.  Ng et al ~\cite{Ng15,DonahueJ2015} process video sequences with CNNs and RNNs, where CNNs are used for frame-level feature extraction and RNNs are used for aggregating video-level temporal information.
 Temporal Segment Networks (TSN) \cite{DBLP:conf/eccv/WangXW0LTG16}, originally designed for 2D CNNs, have been proposed to model video-level spatiotemporal information with video-level sampling strategy, which sample several frames from the whole video during training. The frames are fed to the same 2D CNN backbone, which predicts a confidence score for each frame. The output scores are averaged to generate final video-level prediction. Recently, V4D \cite{Zhang2020V4D:} improves TSN by utilizing 4D convolutions to model short term and long term spatiotemporal features with 3D CNNs. TimeSformer\cite{bertasius2021spacetime} transfers recent vision transformer\cite{dosovitskiy2021an} structures to video domain, surpassing previous clip-level 3D CNNs on various benchmarks.

 {\bf Biases in Spatiotemporal Reasoning Benchmarks}  Biases are hard to be eliminated in machine learning yet can be controlled. For temporal reasoning, Li et al \cite{Li_2018_ECCV} proposed metrics to explicitly calculate the representation biases towards static features in video classification benchmarks. They found that many video benchmarks \cite{DBLP:conf/cvpr/CarreiraZ17,DBLP:journals/corr/abs-1212-0402} lead to high static biases, which makes these benchmarks not suitable for testing the temporal reasoning models. Thus they proposed a new benchmark Diving48, which controls the static biases by using similar foreground and background in all videos thus forcing the model to reason fine-grained temporal structures. CATER also controls the static biases by using simple foreground geometric objects and the same background, requiring the model to understand the whole video to infer which grid the snitch localizes in the last frame. However, we are not using Something-Something dataset in this paper, which is usually considered a temporal-biased benchmark in \cite{Wang_2021_CVPR,Zhang2020V4D:}. We found that static-unbiased and temporal-biased are not the same concept since temporal-biased benchmark can also bias towards static features at the same time. We found Something-Something\cite{DBLP:conf/iccv/GoyalKMMWKHFYMH17} contains strong static biases. Although the benchmark hides the detailed 30408 unique objects categories by a unified term "something", these uncontrolled various foreground entities and backgrounds lead to ambiguity  in terms of comparing temporal reasoning capabilities of different models. Something-Else\cite{Materzynska_2020_CVPR} manually labeled all the foreground objects in something-something and improved I3D baseline by 5\% top1 accuracy with  object identity embeddings.  These phenomenons indicate that something something is also static-biased thus is not a perfect benchmark for temporal reasoning.\\
\begin{figure*}[t]
  \centering
   \includegraphics[width=1\linewidth]{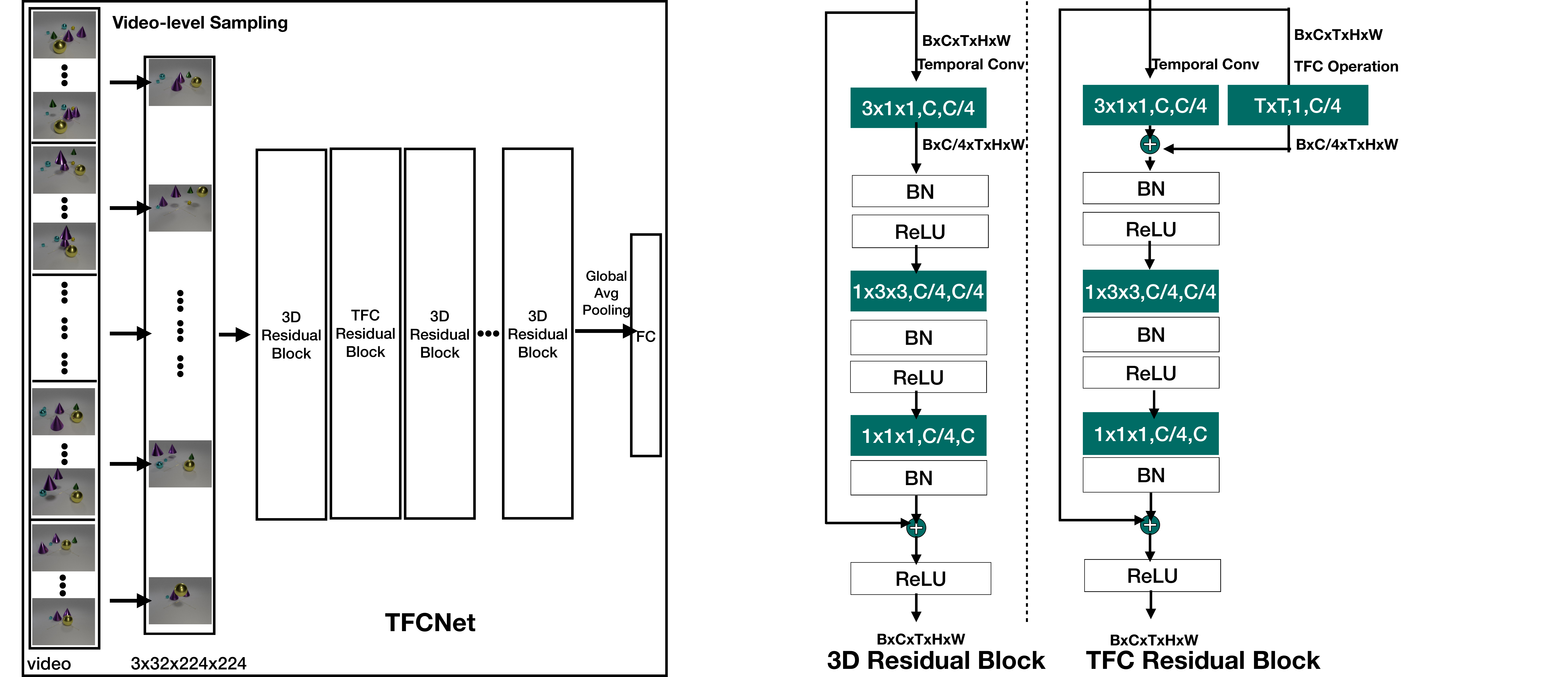}

   \caption{We show the overall framework of TFCNets. Video-level sampling strategy is applied to an input video of any length to form a fixed length input tensor. We replace several 3D Residual Blocks with TFC Residual Blocks to transform V3D to TFCNets. We also demonstrate the detailed structures of 3D Residual Blocks and TFC Residual Blocks with convolution kernels and intermediate features.
   }
   \label{main}
\end{figure*}
\section{Temporal Fully Connected Networks}
\label{sec:TFC}
\subsection{V3D: a new baseline}

Although 3D CNNs are powerful spatiotemporal models, most of them are clip-based methods, which means that only a clip of the whole video will be used for training. In this paper, we choose to use TSN's video-level sampling strategy with 3D CNN structures, so that both the holistic duration can be covered and video-level spatiotemporal features can be learned. The video level sampling strategy is shown in \cref{main}. To be specific,  video-level sampling strategy uniformly divides the whole input video into $T$ segments. During training, one frame is randomly chosen from each segment. During Inference, the center frame of each segment is used. The output of video-level sampling strategy is a tensor of $C\times T \times H \times W$, where $C,T,H,W$ denote input channels, number of frames, height and width of a frame.  We term such {\bf V}ideo-level {\bf 3D} CNNs, V3D, for the ease of reference.

For V3D, we mainly use SlowPath from \cite{DBLP:journals/corr/abs-1812-03982} shown in \cref{tab:arch} for ablation study, which has been proven to be an effective model on various video classification benchmarks. We further propose  V3D Depthwise, where depthwise temporal convolutions are added to 2D TSN ResNet structures. The depthwise convolution is utilized in the residual form \cite{DBLP:conf/cvpr/HeZRS16} with Residual Depthwise Block (RDW) in  \cref{tab:arch}, which is shown in  \cref{eq residual depthwise }. $V_c$ denotes the $c$th channel of input tensor $V$, $O_j$ denotes $j$th dimension of output tensor, with $c=j$ in our case. $W^{depthwise}_{jc}$ is the temporal depthwise convolution. We extend ResNet18 and ResNet50 to V3D Depthwise and we found that V3D Depthwise are able to obtain similar accuracy with V3D, yet with almost the same parameters with 2D TSN and much fewer FLOPs than V3D. 
\begin{equation}
O_j=V_c+W^{depthwise}_{jc} * V_{c}
\label{eq residual depthwise }
\end{equation}
\newcommand{\blockbbasic}[2]{\multirow{2}{*}{\(\left[\begin{array}{c} \text{1\x3\x3, #1}\\[-.1em] \text{1\x3\x3, #1}\end{array}\right]\)\x#2}}
\newcommand{\blocktbasic}[2]{\multirow{4}{*}{\(\left[\begin{array}{c} \text{1\x3\x3, #1}\\[-.1em] \text{RDW 3\x1\x1, #1}\\[- .1em]\text{1\x3\x3, #1}\\\text{RDW 3\x1\x1, #1}\end{array}\right]\)\x#2}}
\newcommand{\blockb}[3]{\multirow{3}{*}{\(\left[\begin{array}{c}\text{1\x1\x1, #2}\\[-.1em] \text{1\x3\x3, #2}\\[-.1em] \text{1\x1\x1, #1}\end{array}\right]\)\x#3}}
\newcommand{\blockt}[3]{\multirow{3}{*}{\(\left[\begin{array}{c}\text{3\x1\x1, #2}\\[-.1em] \text{1\x3\x3, #2}\\[-.1em] \text{1\x1\x1, #1}\end{array}\right]\)\x#3}}
\newcommand{\blockdwt}[3]{\multirow{4}{*}{\(\left[\begin{array}{c}\text{1\x1\x1, #2}\\[-.1em] \text{RDW 3\x1\x1, #2}\\\text{1\x3\x3, #2}\\[-.1em] \text{1\x1\x1, #1}\end{array}\right]\)\x#3}}
\begin{table}[t]
\footnotesize
\centering
\resizebox{0.97\columnwidth}{!}
{
\begin{tabular}{c|c|c|c}
{layer} & TSN ResNet50 & V3D ResNet50& output size \\
\shline
conv$_1$ & \multicolumn{1}{c|}{1\x7\x7, 64, stride 1, 2, 2} &\multicolumn{1}{c|}{1\x7\x7, 64, stride 1, 2, 2} & 32\x112\x112 \\
\hline
\multirow{3}{*}{res$_2$} &\blockb{256}{64}{3}& \blockb{256}{64}{3} & \multirow{3}{*}{32\x56\x56} \\
  &  & \\
  &  & \\
\hline
\multirow{3}{*}{res$_3$} &\blockb{512}{128}{4}& \blockb{512}{128}{4} & \multirow{3}{*}{32\x28\x28} \\
  &  & \\
  &  & \\
\hline
\multirow{3}{*}{res$_4$} &\blockb{1024}{256}{6}& \blockt{1024}{256}{6} & \multirow{3}{*}{32\x14\x14}  \\
  &  & \\
  &  & \\
\hline
\multirow{3}{*}{res$_5$} &\blockb{2048}{512}{3}& \blockt{2048}{512}{3} & \multirow{3}{*}{32\x7\x7} \\
  &  & \\
  &  & \\
\hline
\multicolumn{3}{c|}{3D global average pool, fc} & 1\x1\x1  \\
\hline
\hline
{layer} & V3D Depthwise ResNet18 & V3D Depthwise ResNet50& output size \\
\shline
conv$_1$ & \multicolumn{1}{c|}{1\x7\x7, 64, stride 1, 2, 2} &\multicolumn{1}{c|}{1\x7\x7, 64, stride 1, 2, 2} & 32\x112\x112 \\
\hline
\multirow{4}{*}{res$_2$} &\blocktbasic{64}{2}& \blockb{256}{64}{3} & \multirow{4}{*}{32\x56\x56} \\
  &  & \\
  &  & \\
  & & \\
\hline
\multirow{4}{*}{res$_3$} &\blocktbasic{128}{2}& \blockdwt{512}{128}{4} & \multirow{4}{*}{32\x28\x28} \\
  &  & \\
  &  & \\
  & &\\
\hline
\multirow{4}{*}{res$_4$} &\blocktbasic{256}{2}& \blockdwt{1024}{256}{6} & \multirow{4}{*}{32\x14\x14}  \\
  &  & \\
  &  & \\
  & &\\
\hline
\multirow{4}{*}{res$_5$} &\blocktbasic{512}{2}& \blockdwt{2048}{512}{3} & \multirow{4}{*}{32\x7\x7} \\
  &  & \\
  &  & \\
  & &\\
\hline
\multicolumn{3}{c|}{3D global average pool, fc} & 1\x1\x1  \\

\end{tabular}

}
\vspace{-4mm}
\caption{Baseline: V3D and our proposed V3D Depthwise structures. The example input tensor sampled by video-level sampling strategy is $32\times224\times224$. Please Refer \cref{eq residual depthwise } for detailed RDW structures.
}
\vspace{-1em}
\label{tab:arch}
\end{table}
\subsection{Temporal Fully Connected Operation}
Temporal convolution is a local operator, which has to be stacked multiple times to obtain longer temporal receptive field. In order to obtain video-level global temporal receptive field at any stage of the network, it is a natural idea to utilize fully connected operators to temporal dimension. Yet, fully connected operations easily lead to overfitting with too many parameters and too much computation, thus they are not trivial to be applied to modern CNNs. In this section, we will start from standard temporal convolution, and extend it to temporal fully connected operations, finally reduce the parameters and computation by approximation and simplification.


{\bf Temporal Convolution} Formally, in modern 3D CNNs (which are mixed forms of 2D convolutions and 1D temporal convolutions which are demonstrated in  \cref{tab:arch}), a tensor of an input video is of size $V_{input} \in \mathbb{R}^{B \times C_{in} \times T \times H \times W}$, where $B$ is batch size, $C_{in}$ is number of input channel, $T, H, W$ represent temporal length, spatial height, width respectively. For a standard 1D temporal convolution operation, firstly, the input tensor should be converted to $V \in \mathbb{R}^{BHW \times C_{in} \times T }$ by permutation. Although it can be implemented directly with 3D convolution with a kernel of $3\times1\times1$, we follow the standard 1D temporal convolution notations. Given a temporal 1D convolution kernel $W^{conv} \in \mathbb{R}^{C_{in} \times C_{out} \times K}$, where $C_{in}$ is the number of input channels, $C_{out}$ is the number of output channels, $K$ is the length of temporal kernel, $*$ being convolution operation, a standard temporal convolution can be denoted:
\begin{equation}
O=\cup_j^{C_{out}}\sum\limits_c^{C_{in}}W^{conv}_{jc} * V_{c}
\label{eq temporal conv}
\end{equation}

where $\cup_j^{C_{out}}$ means to concat the $j$th output channel along the $C_{out}$ dimension and $O$ is the output tensor and $O \in \mathbb{R}^{BHW \times C_{out} \times T}$. In recent 3D CNN structures, there is no degradation in temporal dimension where the stride is 1 and paddings are added. Here we do not show such details in the formula for clarity. The FLOPs of temporal convolution in  \cref{eq temporal conv} is $B\times C_{out} \times H \times W \times C_{in} \times T \times K$.\\

{\bf Temporal Fully Connected Operation} It is a simple idea to extend the temporal kernels $W^{conv} \in \mathbb{R}^{C_{in} \times C_{out} \times K}$ to temporal fully connected kernels $W^{connected} \in \mathbb{R}^{C_{in} \times C_{out} \times T \times T}$. Such temporal fully connected kernels increase the temporal modeling length from a local range of $K$  to all frames $T$, where $T$ is the temporal length of input tensor $V$, without degradation in temporal dimension. Formally,
\begin{equation}
O=\cup_j^{C_{out}}\sum\limits_c^{C_{in}}W^{connected}_{jc}V_{c}
\label{eq fully connected conv}
\end{equation}

Also in \cref{eq fully connected conv}, matrix multiplication is used instead of convolutional operation in \cref{eq temporal conv}. Although gaining global receptive field, the parameters of $W^{connected}$ is $T^2/K$ times more than standard temporal convolution. Also the FLOPs of $W^{connected}$ operation is $B\times C_{out} \times H \times W \times C_{in} \times T \times T$, which is $T/K$ times more computation than $W^{conv}$ operation. The parameters and FLOPs make such model easily overfitting and also leads to out of memory issues even on advanced GPUs, which causes fully connected operations rarely used in intermediate stages for modern CNNs.\\

{\bf Simplified TFC Operation} In order to reduce the parameters, we approximate $W^{connected}$ by reducing $C_{in}$ to 1, which means that for each channel of output tensor $o_{j}$, the same matrix $W^{connected}_j \in \mathbb{R}^{1 \times C_{out} \times T \times T} $ is applied for all input channels,
\begin{equation}
O=\cup_j^{C_{out}}\sum\limits_c^{C_{in}}W^{connected}_{j}V_{c}
\label{eq one channel}
\end{equation}
Although the parameters are reduced, the FLOPs is not, still being $B\times C_{out} \times H \times W \times C_{in} \times T \times T$. However, since in \cref{eq one channel} $W^{connected}_{j}$ is the same for all channels, we can exchange the order of channel-dimension summation and  $W^{connected}_{j}$, thus,
\begin{equation}
O=\cup_j^{C_{out}}W^{connected}_{j}\sum\limits_c^{C_{in}}V_{c}
\label{eq one channel converted}
\end{equation}
Since $\sum\limits_c^{C_{in}}V_{c} \in \mathbb{R}^{BHW \times C_{in} \times T}$, the overall FLOPs become $B \times C_{out} \times H \times W \times T \times T$, which decreases the computation cost by $C_{in}$ times. Considering the fact that $C_{in}$ is generally large in modern convolutional neural networks, especially in deeper layers, the computational cost is decreased dramatically. Compared with standard temporal convolution, the FLOPs of equation\cref{eq one channel converted} is $T/(C_{in}K)$ of \cref{eq temporal conv}. Since generally $C_{in}>>T$, our proposed temporal fully connected layer is even much cheaper than standard temporal 1D convolution.

In practice, we find that adding a normalization term along the channel dimension in \cref{eq one channel converted} would make the model converge faster. The normalization term $N(V)=\frac{1}{C_{in}}$ is simply mean average along the channel dimension, making the training more stable. Thus,
\begin{equation}
O=\cup_j^{C_{out}}W^{connected}_{j}\frac{1}{C_{in}}{\sum\limits_c^{C_{in}}V_{c}}
\label{eq one channel normalized}
\end{equation}

We will use {\bf TFC Operation} to refer the Simplified TFC Operation for simplicity in the following sections, unless mentioned otherwise. We provide the Pytorch implementation of TFC Operation \cref{alg:TFC}.  We also compare the parameters and computation of various operations in \cref{cmp_parameter_flops}.

\begin{table*}[!htb]
\scriptsize
\begin{center}
\begin{tabular}{l|l|l}
\hline
{Operation} & {Parameter} & {FLOPs} \\
\hline
\cref{eq temporal conv} & {$C_{out} \times C_{in} \times K$} &{$B \times C_{out} \times H \times W \times T \times C_{in} \times K$}\\
 \hline
\cref{eq fully connected conv} &{$C_{out} \times C_{in} \times T \times T$} & {$B \times C_{out} \times H \times W \times T \times C_{in} \times T$} \\
 \hline
\cref{eq one channel normalized} (ours)&{$C_{out} \times T \times T$} & {$B \times C_{out} \times H \times W \times T \times T$} \\
\hline
SE Operation & {$C_{in} \times C_{in} \times 1/8 $} & {$B \times C_{in} \times C_{in} \times 1/8 $}\\
\hline
Nonlocal Operation & {$C_{in} \times C_{in}/2 \times 4$} & {$B \times C_{in}/2 \times H \times W \times T \times C_{in} \times 4 + B\times
C_{in} \times T \times H \times W \times T \times H \times W 
$}\\
\hline
Temporal Nonlocal Operation & {$C_{in} \times C_{in}/4 \times 3+ C_{in}/4 \times C_{in}/4$} & {$B \times C_{in}/4 \times H \times W \times T \times C_{in} \times 3.25 +
B\times C_{in}/2 \times T \times T \times H \times W 
$}\\

\hline
\end{tabular}
\end{center}
\vspace{-5mm}
\caption{Comparison of Temporal Convolution, TFC Operation and other methods with global temporal receptive field.}
\label{cmp_parameter_flops}
\end{table*}

\begin{algorithm}[t]
\caption{Pseudocode of Temporal Fully Connected Operation in a PyTorch-like style.}
\label{alg:TFC}
\definecolor{codeblue}{rgb}{0.25,0.5,0.5}
\lstset{
  backgroundcolor=\color{white},
  basicstyle=\fontsize{7.2pt}{7.2pt}\ttfamily\selectfont,
  columns=fullflexible,
  breaklines=true,
  captionpos=b,
  commentstyle=\fontsize{7.2pt}{7.2pt}\color{codeblue},
  keywordstyle=\fontsize{7.2pt}{7.2pt},
}
\begin{lstlisting}[language=python]

import torch
class TFC(torch.nn.Module):
    def __init__(self, out_channels=1024,time=32):
        super(TFC, self).__init__()
        #temporal fully connected layer
        self.fc=torch.nn.Linear(time,time*out_channels,bias=False)
        self.c_out=out_channels
    def forward(self,input):
        b,c_in,t,h,w=input.shape
        #permute and view the spatiotemporal tensor
        #to a temporal tensor       
        x=input.permute(0,3,4,1,2).contiguous().view(b*h*w,c_in,t)
        #mean average along input channel dimension
        x=torch.mean(x,dim=1)
        x=x.view(b*h*w,t)
        x=self.fc(x)
        #permute and view the temporal tensor
        #to a spatiotemporal tensor 
        x=x.view(b,h,w,self.c_out,t)
        x=x.permute(0,3,4,1,2).contiguous() 
        #b,c_out,t,h,w
        return x


\end{lstlisting}
\end{algorithm}

\subsection{TFC Residual Block and TFCNet}
With TFC Operation implemented, we change 3D Residual Block  to TFC Residual Block, shown in \cref{main}. For V3D, the TFC Operation is added to the first temporal convolution of 3D Residual Block in a parallel branch, with element-wise addition to merge the local and global temporal features. Similarly, for V3D Depthwise, we add the TFC Operation in parallel with the first 2D convolution kernel of the Residual Block. We turn every other 3D Residual Blocks in res3 and res4 to TFC Residual Blocks.  For 2D TSN baseline, we  add TFC Operations to the first 2D kernel of every Residual Blocks in res4 and res5. We term these baseline networks with TFC Blocks "{\bf TFCNets}".

\subsection{Comparison with previous methods with global receptive field}
We notice that there are some previous operations also able to obtain global receptive field. Two typical such methods are SE Block and Non-local Block. Here we will discuss and compare our proposed TFC layer with these methods.\\
{\bf SE Block } Hu et al \cite{Hu_2018_CVPR} proposed SE Block for 2D image classification. We can simply extend SE Block to its 3D version, which is able to model spatiotemporal features.  The spatiotemporal SE block adopts 3D global average pooling on the whole input tensor $V \in \mathbb{R}^{B \times C_{in} \times T \times H \times W}$ to squeeze the $T,H,W$ dimensions into one scalar, in which way the global view is obtained. Two fully-connected layers and ReLU activations are further adopted to generate the channel-wise attention maps, followed by expanding the $B \times C_{in}$ attention maps to $B \times C_{in} \times T \times H \times W$ and element-wise multiplication with the input tensor $V$.

Compared with our TFC Operation, SE Block obtains global receptive field by squeezing spatiotemporal dimensions into one scalar, which is efficient, yet completely loses fine-grained spatiotemporal structures. Also it has to be used in each Residual Block according to the authors to obtain best accuracy, which leads to more overall parameters than TFCNet. We will show in our experiments that 3D SE Block actually hurts the performance of 3D CNNs on long-range video reasoning. Instead, our TFC Operation does not make any dimension degradation, neither spatial dimensions nor temporal dimension.

{\bf Non-local Blocks} Wang et al \cite{DBLP:conf/cvpr/0004GGH18}proposed Nonlocal Blocks to model global spatiotemporal features, which can be considered one typical implementation of self-attention methods. Nonlocal blocks generates attention maps of $THW \times THW$ by applying matrix multiplication on two tensors of size $THW \times C/2$. The computation cost of Nonlocal Blocks or self attention methods are composed of two parts, the embedded layers and matrix multiplication. We notice that the embedded layers have already lead to  more computation than our proposed TFC Operation.  If spatiotemporal attention is calculated, the computation cost of Nonlocal Blocks becomes hundreds times of our TFC Operation and Nonlocal Blocks cannot be trained with the settings of TFCNets due to out of memory issue. Although we use linear scaling rule \cite{DBLP:journals/corr/GoyalDGNWKTJH17} to adjust the learning rate, spatiotemporal Nonlocal V3D with 32 frames in temporal dimension cannot converge with too few samples in a batch. It is possible to ease this problem with other techniques, yet this would completely change the settings and lead to unfair comparison. Instead, we use temporal Nonlocal Blocks to compare with our TFC Blocks. We change the input and output channels of four 1x1 convolutions in the temporal Nonlocal Block so that our TFC Operation in a 3D Residual Block can be directly replaced by the temporal Nonlocal Block.

\subsection{Limitation}
Our proposed TFC Operation comes with one inevitable disadvantage, which is the fixed length of temporal dimension. This means that the number of input frames during training must be identical with the frames during inference. Yet this is not a problem for classification, which is the task of this paper, since the video-level sampling strategy perfectly solves this issue by its variant sampling intervals. We will show in the experiment section that TFCNets achieve new state-of-the-art accuracy on Diving48 benchmark, whose input videos are of different length. However, for downstream tasks which use TFCNets to extract features, it remains to be an open problem, which we will explore in the future.
\section{Experiment}
\subsection{Dataset}
We use two static unbiased temporal reasoning benchmarks to demonstrate the strong temporal reasoning ability of TFCNet. CATER \cite{Girdhar2020CATER:} is a synthetic dataset, where videos use the same background with simple geometry objects moving  in the foreground. The most challenging task in CATER is the snitch localization, shown in \cref{fig:intro}, which requires the temporal model understanding object permanence. There are 3850 videos for training and 1650 videos for validation. There are 36 classes, corresponding to the $6\times6$ grid. The videos are all of 300 frames.  We also validate TFCNets on real world temporal reasoning benchmark, Diving48 \cite{Li_2018_ECCV}. There are 48 fine-grained categories with 16k videos for training, 2k for testing. Since there is no official validation set, the testing set is used for validation by all previous methods. We follow the same criterion for fair comparison. The video lengths are variant, with 158 frames on average,  min 24 frames and max 822 frames.

\subsection{implementation details}
We use pre-trained weights from ImageNet-1k \cite{DBLP:conf/cvpr/DengDSLL009} to initialize the model.
For  TSN, V3D baselines and our proposed TFCNets, we adopt video-level sampling strategies where we uniformly sample the same number of frames covering the whole video for both training and testing. We first resize each frame to $320 \times 240$. The only data augmentation we used are simple random cropping  of $224 \times 224$ and random crop. 
 During testing, we use center cropping to crop a region of $224 \times 224$ at the center of each frame. We keep the training and inference pipeline simple and effective, without any tricks.

In order to demonstrate the importance of video-level sampling strategy,  we also experiment with the 3D CNN and TFCNet structures with clip-level sampling strategy. For the clip-level sampling, following \cite{DBLP:journals/corr/abs-1812-03982}, we first resize each frame to $320 \times 256$. During training, we randomly select a clip of 64 frames from each video where we uniformly sample 32 frames in this clip with a fixed stride of 2. During Testing we use spatial fully convolutional testing by following \cite{DBLP:conf/cvpr/0004GGH18,DBLP:conf/nips/YueSYZDX18,DBLP:journals/corr/abs-1812-03982}. We
sample 10 clips evenly from the full-length video, and crop  $256 \times 256$ regions to spatially cover the whole frame for each clip. 

 We utilize a SGD optimizer with an initial learning rate of 0.01, weight decay is set to $10^{-5}$ with a momentum of 0.9. The batch size is 8 on each GPU, with 8 GPUs the total batch size is 64. The learning rate drops by 10 at epoch 35, 65, and the model is trained for 90 epochs in total.
\begin{table*}[t]\centering
\subfloat[Effectiveness of TFCNet.\label{effectiveness}]{
\begin{tabular}{c|c|c|c|c|c|c|c|c}
\hline
{model} &{backbone} & {input $T \times H \times W$} & {top1} & {gain}&{top5}&{L1 Loss} &{parameters} &{GFLOPs}\\
\hline
{TSN} & ResNet50 & {$32\times224\times224$} & 34.1&- & 57.0 &2.33&23.58M&131.5 \\
TFC TSN & ResNet50 & {$32\times224\times224$}& 73.6 &{+39.5\%}& 92.3 &0.68 &26.65M&132.1\\
\hline
V3D & {ResNet50} & {$32\times224\times224$} & 77.9 &-& 94.0 & 0.54 &31.63M &167.7\\
TFC V3D &{ResNet50} & {$32\times224\times224$} & 79.5 &{+1.6\%}& 94.0 & 0.50 &32.68M &168.0 \\

\hline
V3D Depthwise & {ResNet18} & {$32\times224\times224$} &76.1 &-&93.5&0.57&11.19M&58.3\\
{TFC V3D Depthwise}&{ResNet18} & {$32\times224\times224$} & 77.4& +1.3\%&94.3& 0.52&11.97M&58.6\\
\hline
{V3D Depthwise}&{ResNet50} & {$32\times224\times224$} &76.6 &-&94.6&0.51&23.59M&131.6\\
{TFC V3D Depthwise}&{ResNet50} & {$32\times224\times224$} & 79.7& +3.1\%&95.5 & 0.47&24.64M&132.0\\
\hline
\end{tabular}}\hspace{3mm}
\subfloat[video-level sampling. \label{sampling}]{
\begin{tabular}{c|c|c|c|c|c|c}
\hline
{model}   &{sampling strategy} & {top1} & {top5} & {frame $\times$ segment $\times$ crops} & {parameters} & {GFLOPs}\\
\hline
{V3D ResNet50} & {clip} & 23.1 & 46.4 &{$32 \times 10 \times 3$} &31.63M & 6571.1\\
{V3D ResNet50} & {video}  & 77.9 & 94.0 &{$1 \times 32 \times 1$}&31.63M&167.7\\
\hline
{V4D ResNet50\cite{Zhang2020V4D:}} & {clip-video} & 43.4 & 63.1&{$8 \times 4 \times 1$} & 36.42M & 373.6\\
{V4D ResNet50\cite{Zhang2020V4D:}} & {clip-video} & 47.2 & 72.1 & {$8  \times 10 \times 3$} & 36.42M & 934.0\\

\hline
{TFC V3D ResNet50} & {clip} & 25.2 & 54.0 &{$32 \times 10 \times 3$}&32.68M & 6582.9\\
{TFC V3D ResNet50} & {video}  & 79.5 & 94.0 &{$1 \times 32 \times 1$}&32.68M & 168.0\\

\hline
\end{tabular}}\hspace{3mm}
\subfloat[Different input length\label{num of frame}]{
\begin{tabular}{c|c|c|c|c}
\hline
{model}  & {training and inference segment} & {top1} & {paramters}&{GFLOPs}\\
\hline
{V3D ResNet50} & 8 & 55.2 & 31.63M & 41.9\\
{TFC V3D ResNet50} & 8 &54.6 & 31.70M & 41.9\\
\hline
{V3D ResNet50} & 16  & 69.7 & 31.63M & 83.8 \\
{TFC V3D ResNet50} & 16& 70.2&  31.90M & 83.9\\
\hline
{V3D ResNet50} & 32 & 77.9 &  31.63M & 167.7\\
{TFC V3D ResNet50} & 32 & 79.5 & 32.68M&168.0\\
\hline
\end{tabular}}\hspace{3mm}
\subfloat[compare against SE Network and Nonlocal Network \label{senet nonlocal}]{
\begin{tabular}{c|c|c|c|c}
\hline
{model}  & {input size}&{top1}& {paramters}&{GFLOPs}\\
\hline
{V3D ResNet50} &{$32 \times 224 \times 224$} & 77.9  &31.63M &167.7\\
{ SE V3D ResNet50} & {$32 \times 224 \times 224$}  & 77.0  &34.14M &167.9 \\
{ Non-local V3D ResNet50} & {$32 \times 224 \times 224$} & 78.6&34.62M &195.4\\
\hline
{TFC V3D ResNet50} & {$32 \times 224 \times 224$} & 79.5 & 32.68M &168.0\\

\hline
\end{tabular}}



\vspace{-4mm}
\caption{\textbf{Ablations} on CATER snitch localization.}
\label{tab:mini-kinetics}
\end{table*}
\begin{table*}[!htb]
\tablestyle{3pt}{1}
\begin{center}
\begin{tabular}{c|c|c|c|c|c|c|c|c}
\hline
{Model} & {Backbone} & {total frames } &{ extra annotation/data} & {top1} & {top5} & {L1 Loss} & {params} & {GFLOPs}\\
\hline
Tracking\cite{zhu2018distractoraware} & {SiamRPN} & {all frames} & {-}& 33.9 & -&2.4&-&- \\
R3D\cite{DBLP:conf/cvpr/0004GGH18}  &{ResNet50} & {$64 \times 10 \times 3 $} &{-} &57.4 & 78.4 &1.4&-&-\\
R3D+LSTM\cite{Girdhar2020CATER:}&{ResNet50}&{$64 \times 10 \times 1 $}&-&60.2 &81.8& 1.2&-&-\\
R3D+NL\cite{DBLP:conf/cvpr/0004GGH18}&{ResNet50} & {$32 \times 10 \times 3$} &-&26.7 & 68.9 & 2.6&-&-\\
R3D+NL+LSTM\cite{Girdhar2020CATER:}  &{ResNet50} & {$32 \times 10 \times 1$} &- &46.2& 69.9& 1.5&-&-\\
Hopper\cite{zhou2021hopper}  & {Transformer} & {all frames } &{LA-CATER\cite{shamsian2020learning}} & 73.2 & 93.8 & 0.85&-&-\\
OPNet\cite{shamsian2020learning} & {OPNet} & {all frames} &{LA-CATER\cite{shamsian2020learning}}&74.8 & - & 0.54&-&-\\

V4D \cite{Zhang2020V4D:}&{ResNet50} & {$8 \times 10 \times 3$} & {-} & 47.2 &  72.1 & -&36.42M &934.0\\
\hline
V3D & {ResNet50} & {$1\times 32 \times 1$}&-&77.9&94.0&0.54 &31.63M & 167.7  \\
TFC V3D (ours)&{ResNet50} & {$1\times 32 \times 1$}&-&79.5&94.0&0.50&32.68M&168.0\\
TFC V3D Depthwise (ours) & {ResNet50} & {$1\times 32 \times 1$}&- & {\bf 79.7} &{\bf 95.5} & {\bf 0.47}&24.64M&132.0\\
\hline
\end{tabular}
\end{center}
\vspace{-5mm}
\caption{Comparison with state-of-the-art on CATER snitch localization.}
\label{cmp_miniart}
\end{table*}
\begin{table*}[!htb]
\begin{center}
\begin{tabular}{c|c|c|c|c|c|c}
\hline
{Model} & {Backbone} & {segment $\times$ crop }  & {top1} & {top5} &  {parameters} & {GFLOPs}\\
\hline
SlowFast \cite{DBLP:journals/corr/abs-1812-03982}& {ResNet101} & {$128 \times 3$} &77.6& -& 53.7M &213x3 \\
TimeSformer-L\cite{bertasius2021spacetime} & {TimeSformer}  & {$96 \times 3$}  &81.0 &-&121.4M &2380x3\\
TQN\cite{zhangtqn}&{S3D+Feature Bank}&all frames&81.8&- &- &-\\
\hline

TFC V3D Depthwise (ours) & {ResNet50} & {$32\times 1$} & 87.9 &98.2 &24.65M&132.1\\
TFC V3D Depthwise (ours) & {ResNet50} & {$32\times 3$} & 88.3 & 98.3&24.65M&132.1x3\\
\hline
\end{tabular}
\end{center}
\vspace{-5mm}
\caption{Results on Diving48 V2.}
\label{diving48 v2}
\end{table*}
\subsection{CATER}
We thoroughly explore the properties of our proposed TFCNet by conducting ablation studies on CATER snitch localization\cite{Girdhar2020CATER:}. Besides Top1 accuracy, L1 Loss is also used for the metrics following \cite{Girdhar2020CATER:}.

{\bf Effectiveness of TFCNet} We compare against 4 types of video-level baselines, Temporal Segment Networks, V3D ResNet50, V3D Depthwise ResNet18 and V3D Depthwise ResNet50. In order to compare the structures fairly, all of the input are of size $3 \times 32 \times 224 \times 224$, where 3 is the RGB channel, 32 is the number of frames, $224 \times 224$ is the center crop of each frame. 

We show in \cref{effectiveness} that TFCNet consistently improve various backbones with very few extra parameters and GFLOPs. For TSN ResNet50, TFCNet improves top1 accuracy of TSN by $39.5\%$ with 3.1M more parameters and 0.6GFLOPs more computation, proving the importance of learning long-range temporal interaction for spatiotemporal reasoning. For V3D ResNet50, please note that our V3D ResNet50 baseline has already surpassed all previously published state-of-the-art results on CATER\cite{Girdhar2020CATER:}, which indicates that V3D is a very strong baseline for static-unbiased learning. Our TFC Blocks further improve V3D by $1.6\%$, enabling global temporal receptive field in the intermediate stages with only 1M extra parameters and 0.3GFLOPs more computation.  Our TFC V3D Depthwise ResNet50 further establish new state-of-the-art results  on CATER snitch localization benchmark.

{\bf video-level vs clip-level } Most previous 3D CNNs are clip-based models, which means that during training only a short clip from the whole video is randomly sampled. For long-range spatiotemporal reasoning tasks, e.g. CATER, where the spatiotemporal patterns change from the first frame to the last frame, it is obvious that such local sampling strategy will not work well. We compare clip-level sampling strategy against video-level ones to demonstrate the importance of video-level learning. Shown in \cref{sampling}, for both V3D baselines and our proposed TFC V3D, video-level sampling strategy leads to consistent improvements of about $54\%$ over the clip-level sampling strategy. Also the standard clip-level methods use much more clips and higher resolution (224 to 256), leading to 39 times more computation cost than video-level counterparts.

Besides, \cref{sampling} also proves that our proposed TFC Block is a general module that not only improves video-level methods, but also is  able to lead to $2.1\%$ improvement against 3D baseline even in the clip-level learning setting. Finally, we also compare the video-level sampling strategy against the recently proposed V4D\cite{Zhang2020V4D:}, which adopts hybrid sampling strategy that is locally clip-level and globally video-level. We demonstrate that video-level sampling strategy still outperforms V4D sampling strategy by more than $30\%$ absolutely. However, our experiment shows that V4D leads to more than $20\%$ absolute improvement over clip-level 3D CNNs, which is consistent with the conclusion of V4D.

{\bf Different temporal length} We compare the improvement of TFC Blocks with different input frames for training and inference. Please note that the parameters of Temporal Fully Connected Operations are quadratic to the temporal length, thus the parameters of TFC V3D with different input frames are slightly different, shown in \cref{num of frame}. We found that when the input frames are rather sparse, for example, 8 frames, adding TFC Block causes decrease in top1 accuracy. With more frames used for training, the advantage of TFCNet becomes more significant. We infer that this phenomenon is due to the limited temporal receptive field of V3D. When the temporal length is only 8 frames, the receptive field of V3D is enough to learn temporal interactions. However, with input frame increased to 32, it becomes harder for V3D to learn global spatiotemporal representation. Our proposed TFC Blocks make it possible to obtain global temporal receptive field at intermediate stages for V3D, thus lead to higher accuracy.

{\bf Comparison against SE Block and Nonlocal Block} SE Block and Nonlocal Block are another two typical structures to obtain global receptive field in modern neural networks. We compare our proposed TFC Block with SE Block and Nonlocal Block from the aspects of accuracy, parameters and GFLOPs. We extend the original 2D SE Block to 3D spatiotemporal SE Block and we follow \cite{Hu_2018_CVPR} to insert one SE Block to every Residual Bottleneck for best performance. For Nonlocal Blocks, we found the original spatiotemporal Nonlocal Blocks\cite{DBLP:conf/cvpr/0004GGH18} lead to out of memory issues with our V3D structures, thus we experiment with temporal Nonlocal Blocks, replacing the TFC Operation branches in 3D Residual Blocks for fair comparison. For original Nonlocal Networks \cite{DBLP:conf/cvpr/0004GGH18}, we also compare their results with TFCNets on CATER in the following section.

In \cref{senet nonlocal}, we show that compared to baseline V3D ResNet50, SE V3D ResNet50 actually causes decrease in accuracy. We believe this is because SE Block applies spatiotemporal global average pooling to obtain global receptive field, with the side effect of destroying fine-grained spatiotemporal structures. This phenomenon indicates the importance of keeping the dimensions of space and time for the intermediate features.

Also, we found that Nonlocal Blocks do increase the accuracy of  V3D network. However, Nonlocal V3D leads to 27GFLOPs more computation cost than V3D, compared with our TFC V3D with only 0.3GFLOPs more computation. Also, our TFC V3D has the highest accuracy among these modules of global receptive field and is the cheapest in terms of parameters.

 These experiments also prove that the accuracy gain of TFCNets is {\bf{\bf NOT} simply due to adding more parameters.}

{\bf Compare with State-of-the-art} We compare our proposed TFCNets with previous state-of-the-art results reported on this dataset, including various CNNs, CNN+RNN, and recent vision transformers. Our TFCNets establish new state-of-the-art results on this benchmark, surpassing all previous methods, including the ones utilizing external training data or annotations\cite{shamsian2020learning}.

\subsection{Diving48}
We also test the temporal reasoning ability of TFCNet on real world dataset, Diving48. The top1 accuracy of TFCNet on Diving48 V2 is shown in \cref{diving48 v2}. With the fewest parameters and FLOPs, our TFCNets achieve new state-of-the-art top1 accuracy of 88.3\%. Compared with recent video transformers, our TFCNets are  7.3\% higher in terms of accuracy. For Diving48 V1, since there are many videos mislabeled, it does not accurately reflect the temporal reasoning ability of different models. Here we report the results of TFCNet on Diving48 V1 only for completeness. Compared to the best published result 44.7\% top1 accuracy from CorrNet ResNet101\cite{wang2020video}, our TFCNet ResNet50 achieves 56.3\% top1 accuracy with 11.6\% absolute increase.

\section{Conclusion}
In this paper, we study static unbiased temporal reasoning in videos and propose Temporal Fully Connected Networks to learn video-level spatiotemporal features. We propose very efficient and effective TFC Operations, combined with V3D and our proposed depthwise variants. TFCNets achieve new state-of-the-art results on object permanence dataset CATER and real world temporal reasoning benchmark Diving48. 


\bibliography{example_paper}
\bibliographystyle{icml2022}


\end{document}